\documentclass{article}

 \usepackage[final]{neurips_2021}
 \usepackage{multirow}
 \usepackage{wrapfig}
 \usepackage{multicol}
 \usepackage{paralist}

\usepackage[utf8]{inputenc} 
\usepackage[T1]{fontenc}    
\usepackage{url}            
\usepackage{booktabs}       
\usepackage{amsfonts}       
\usepackage{nicefrac}       
\usepackage{microtype}      
\usepackage{xcolor}         
\usepackage{amsmath}
\usepackage{subcaption}

\usepackage{graphicx}
\usepackage{float}
\usepackage{algorithmicx}
\usepackage[ruled]{algorithm}
\usepackage{algpseudocode}
\usepackage{hyperref}

\def\our{NGGP}
\def\GP{\mathcal{GP}}

\title{
 Non-Gaussian Gaussian Processes\\for Few-Shot Regression
}

\author{%
  Marcin Sendera \thanks{Corresponding author: \texttt{marcin.sendera@doctoral.uj.edu.pl}}\\ 
  Jagiellonian University\\
  \And
  Jacek Tabor\\
  Jagiellonian University\\
  \And
  Aleksandra Nowak\\
  Jagiellonian University\\
  \And
  Andrzej Bedychaj\\
  Jagiellonian University\\
  \And
  Massimiliano Patacchiola\\
  University of Cambridge\\
  \newline
  \AND
  Tomasz Trzcinski\\
  Jagiellonian University, \\
  Warsaw University of Technology, \\
  Tooploox\\
  \And
  Przemysław Spurek\\
  Jagiellonian University\\
  \And
  Maciej Zieba\\
  Wrocław University of \\Science and Technology,\\
  Tooploox\\
}

\usepackage[compact]{titlesec}
\usepackage{paralist}
\titlespacing{\section}{0.2pt}{0.5ex}{0ex}
\titlespacing{\subsection}{0.1pt}{0.2ex}{0ex}
\titlespacing{\paragraph}{0.1pt}{0ex}{1ex}

\begin{document}

\maketitle

\begin{abstract}

Gaussian Processes (GPs) have been widely used in machine learning to model distributions over functions, with applications including multi-modal regression, time-series prediction, and few-shot learning. GPs are particularly useful in the last application since they rely on Normal distributions and enable closed-form computation of the posterior probability function. Unfortunately, because the resulting posterior is not flexible enough to capture complex distributions, GPs assume high similarity between subsequent tasks -- a~requirement rarely met in real-world conditions.
In this work, we address this limitation by leveraging the flexibility of Normalizing Flows to modulate the posterior predictive distribution of the GP. This makes the GP posterior locally non-Gaussian, therefore we name our method Non-Gaussian Gaussian Processes (NGGPs). 
We propose an invertible ODE-based mapping that operates on each component of the random variable vectors and shares the parameters across all of them. 
We empirically tested the flexibility of NGGPs on various few-shot learning regression datasets, showing that the mapping can incorporate context embedding information to model different noise levels for periodic functions.
As a result, our method shares the structure of the problem between subsequent tasks, but the contextualization allows for adaptation to dissimilarities.
NGGPs outperform the competing state-of-the-art approaches on a diversified set of benchmarks and applications.
\end{abstract}

\section{Introduction}

Gaussian Processes (GPs)~\cite{rasmussen2003gaussian,trapp2020deep} are one of the most important probabilistic methods, and they have been widely used to model distributions over functions in a variety of applications such as multi-modal regression~\cite{zikeba2020regflow}, time-series prediction~\cite{brahim2004gaussian,najibi2021enhanced} and meta-learning~\cite{patacchiola2020bayesian,tossou2019adaptive}. Recent works propose to use GPs in the {\em few-shot learning} scenario ~\cite{chen2019closer,patacchiola2020bayesian,snell2020bayesian,wang2020generalizing}, where the model is trained to solve a supervised task with only a few labeled samples available. This particular application is well-fitted to GPs since they can determine the posterior distribution in closed-form from a small set of data samples~\cite{patacchiola2020bayesian}.

However, the generalization capabilities of GPs come at the price of reduced flexibility when the modeled distributions are complex, {\it e.g.}, they have high skewness or heavy tails. Furthermore, GPs assume a high similarity between subsequent tasks. This condition is rarely met in real-world applications where tasks can vary during time, as is the case in heteroscedastic regression. These limitations of GPs also extend to multi-modal learning or, more generally, to multi-label regression~\cite{zikeba2020regflow}.

\begin{figure}[t]
  \begin{center}
    \includegraphics[width=0.4\textwidth]{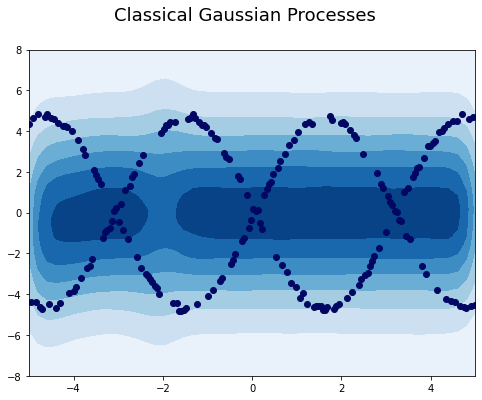}
    \includegraphics[width=0.4\textwidth]{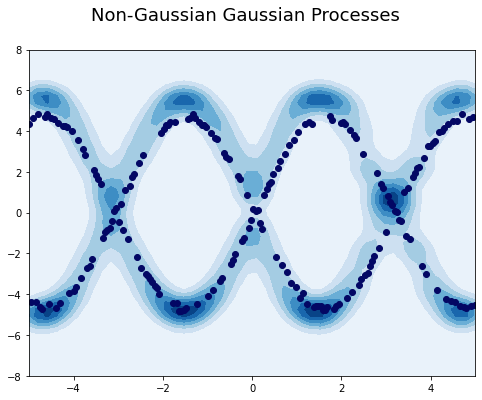}
  \end{center}
  \vspace{-0.3cm}
  \caption{Results of Deep Kernels with classical GP (left) and \our{} (right). The one-dimensional samples were generated randomly from $\sin(x)$ and $-\sin(x)$ functions with additional noise. \our{}, compared to GP,  does not have an assumption of Gaussian prior, which allows for modeling a multi-modal distribution. }\label{fig:tezer}
\end{figure}

In this work, we address those drawbacks by modeling the GPs posterior predictive distributions with a local non-Gaussian approximation. We do so by introducing a new method that we have named \emph{Non-Gaussian Gaussian Processes (NGGPs)}. In NGGPs, we leverage the flexibility of Continuous Normalizing Flows (CNF)~\cite{grathwohl2018ffjord} to model arbitrary probability distributions. In particular, we propose an invertible ODE-based mapping that operates on each component of the random variable vectors. This way, we can compute a set of CNFs parameters shared across all vectors, with the resulting mapping incorporating the information of the context to model different noise for periodic functions.
Figure~\ref{fig:tezer} shows how NGGPs are able to capture the overall structure of a problem, whereas standard GPs fail. NGGPs are able to reconstruct a multi-modal sine function while adapting to local dissimilarities thanks to the contextualization provided by the ODE-based mapping. We provide empirical evidence that NGGPs outperform competitive state-of-the-art approaches on a diversified set of benchmarks and applications in a few-shot learning scenario; the code is released with an open-source license\footnote{\url{https://github.com/gmum/non-gaussian-gaussian-processes}}.

The contributions of our work can be summarized as follows:
\begin{compactitem}
    \item We introduce Non-Gaussian Gaussian Processes (NGGPs), a new probabilistic method for modeling complex distributions through locally non-Gaussian posteriors.
    \item We show how invertible ODE-based mappings can be coupled with GPs to process the marginals of multivariate random variables resulting in more flexible models.
    \item We extensively test NGGPs on a variety of few-shot learning benchmarks, achieving state-of-the-art performances in most conditions.
\end{compactitem}

\section{Related Work}

The related work section is divided into three parts. First, we present a general Few-Shot Learning problem. Then, we discuss GPs, focusing on models, which use flow architectures. Finally, in the third paragraph, we describe existing approaches to Few-Shot Learning, which use Gaussian Processes.

\paragraph{Few-Shot Learning}
Few-Shot Learning aims at solving problems in which the number of observations is limited. Some of the early methods in this domain have applied a two-phase approach by pre-training on the base set of training tasks and then fine-tuning the parameters to the test tasks~\cite{chen2019closer,pan2009survey}. An alternative approach is given by non-parametric metric-learning algorithms, which aim at optimizing a metric, that is then used to calculate the distance between the target observations and the support set items~\cite{vinyals2016matching, snell2017prototypical, sung2018learning}. 
Another popular approach to few-shot learning is Model Agnostic Meta-Learning (MAML) \cite{finn2017model} and its variants \cite{garnelo2018neural,li2020international,rajeswaran2019meta,yin2019meta, gordon2018meta, xu2020metafun, du2021metakernel}. MAML aims at finding a set of joined task parameters that can be easily fine-tuned to new test tasks via few gradient descent updates. MAML can also be treated as a Bayesian hierarchical model ~\cite{finn2018probabilistic,grant2018recasting,jerfel2018reconciling}. Bayesian MAML \cite{yoon2018bayesian} combines efficient gradient-based meta-learning with non-parametric variational inference in a principled probabilistic framework. 
A few algorithms have been focusing exclusively on regression tasks. An example is given by ALPaCA \cite{harrison2018meta}, which uses a dataset of sample functions to learn a domain-specific encoding and prior over weights. 

\paragraph{Gaussian Processes}
GPs have been applied to numerous machine learning problems, such as spatio-temporal density estimation  \cite{dutordoir2018gaussian}, robotic control \cite{yeung2009learning}, or dynamics modeling in transcriptional processes in the human cell \cite{NIPS2006_f42c7f9c}.
The drawback of GP lies in the computational cost of the training step, which is $O(n^3)$ (where $n$ denotes the number of observations in the training sample).

In \cite{snelson2004warped}, the authors extend the flexibility of GPs by processing the targets with a learnable monotonic mapping (the warping function). This idea is further extended in \cite{lazaro2012bayesian}, which shows that it is possible to place the prior of another GP on the warping function itself. Our method is different from these approaches, since the likelihood transformation is obtained by the use of a learnable CNF mapping.

In \cite{maronas2021transforming}, the authors present the Transformed Gaussian Processes (TGP), a new flexible family of function priors that use GPs and flow models. TGPs exploit Bayesian Neural Networks (BNNs) as input-dependent parametric transformations. The method can match the performance of Deep GPs at a fraction of the computational cost.  

The methods discussed above are trained on a single dataset, that is kept unchanged. Therefore, it is not trivial to adapt such methods to the the few-shot setting.

\paragraph{Few-Shot Learning with Gaussian Processes}
When the number of observations is relatively small, GPs represent an interesting alternative to other regression approaches. This makes GPs a good candidate for meta-learning and few-shot learning, as shown by recent publications that have explored this research direction. For instance, Adaptive Deep Kernel Learning (ADKL) \cite{tossou2019adaptive} proposes a variant of kernel learning for GPs, which aims at finding appropriate kernels for each task during inference by using a meta-learning approach. 
A similar approach can be used to learn the mean function \cite{fortuin2019meta}.
In \cite{rothfuss2021pacoh}, the authors presented a theoretically principled PAC-Bayesian framework for meta-learning. It can be used with different base learners (e.g., GPs or BNNs). 
Topics related to kernel tricks and meta-learning have been explored in \cite{venkitaraman2020task}. The authors propose to use nonparametric kernel regression for the inner loop update. 
In \cite{titsias2020information}, the authors introduce an information-theoretic framework for meta-learning by using a variational approximation to the information bottleneck. In their GP-based approach, to account for likelihoods other than Gaussians, they propose approximating the non-Gaussian terms in the posterior with Gaussian distributions (by using amortized functions), while we use CNFs to increase the flexibility of the GPs.

In \cite{patacchiola2020bayesian}, the authors present Deep Kernel Transfer (DKT): a Bayesian treatment for the meta-learning inner loop through the use of deep kernels, which has achieved state-of-the-art results.  
In DKT, the deep kernel and the parameters of the GP are shared across all tasks and adjusted to maximize the marginal log-likelihood, which is equivalent to Maximum-Likelihood type II (ML-II) learning. DKT is particularly effective in the regression case since it is able to capture prior knowledge about the data through the GP kernel. However, in many settings, prior assumptions could be detrimental if they are not met during the evaluation phase. This is the case in few-shot regression, where there can be a significant difference between the tasks seen at training time and the tasks seen at evaluation time. For instance, if we are given few-shot tasks consisting of samples from periodic functions but periodicity is violated at evaluation time, then methods like DKT may suffer in terms of predictive accuracy under this domain shift. In this work, we tackle this problem by exploiting the flexibility of CNFs. 

\section{Background}
\label{sec:bg}

\paragraph{Gaussian Processes.} The method proposed in this paper strongly relies on Gaussian Processes (GPs) and their applications in regression problems. GPs are a well-established framework for principled uncertainty quantification and automatic selection of hyperparameters through a marginal likelihood objective~\cite{GP_rasmussen}. More formally, a GP is a collection of random variables such that the joint distribution of every finite subset of random variables from this collection is a multivariate Gaussian~\cite{quinonero2005unifying}. We denote Gaussian Process as $f(\cdot) \sim \GP(\mu(\cdot),k(\cdot,\cdot))$, where $\mu(\mathbf{x})$ and $k(\mathbf{x},\mathbf{x'})$ are the mean and covariance functions.
When prior information is not available, a common choice for $\mu$ is the zero constant function. The covariance function must impose a valid covariance matrix. This is achieved by restricting $k$ to be a kernel function.
Examples of such kernels include the Linear kernel, Radial Basis Function (RBF) kernel, Spectral Mixture (Spectral) kernel~\cite{wilson2013gaussian},  or Cosine-Similarity kernel~\cite{rasmussen2003gaussian}. Kernel functions can also be directly modeled as inner products defined in the feature space imposed by a feature mapping  $\psi: X \rightarrow V$: 
\begin{equation}
k(x,x') = \langle\psi(x),\psi(x')\rangle_{V}     
\label{eq:kernel}
\end{equation}
An advantage of the formulation above is that it can be easily implemented by modeling $\psi$ through a neural network. 
Throughout this work, we call this technique the NN Linear kernel (sometimes called Deep Kernel ~\cite{patacchiola2020bayesian}). Since every kernel can be described in terms of Equation~\eqref{eq:kernel}, such an approach may be desired if no prior information about the structure of the kernel function is available. 

Gaussian Processes provide a method for modeling probability distributions over functions. 
Consider a regression problem:
\begin{equation}
 \mathbf{y}_i=f(\mathbf{x}_i)+\epsilon_i, \textrm{ for } i=1,\ldots,m,
\end{equation}
where $\epsilon_i$ are i.i.d. noise variables with independent $\mathcal{N}(0,\sigma^2)$ distributions. 
Let $\mathbf{X}$ be the matrix composed of all samples $\mathbf{x}_i$ and let $\mathbf{y}$ be the vector composed of all target values $\mathbf{y}_i$. 
Assuming that $f(\cdot) \sim \GP\left(0,k\left(\cdot,\cdot\right)\right)$, we obtain:
\begin{equation}
\mathbf{y}|\mathbf{X} \sim \mathcal{N}(0,\mathbf{K}+\sigma\mathbb{I}),
\label{eq:GP_train}
\end{equation}
where $k_{i,j} = k(\mathbf{x}_i,\mathbf{x}_j)$. Analogously, inference over the unknown during the training samples is obtained by conditioning over the normal distribution. Let $(\mathbf{y},\mathbf{X})$ be the train data and let $(\mathbf{y}_*, \mathbf{X}_*)$ be the test data. Then the distribution of $\mathbf{y}_*$ given $\mathbf{y},\mathbf{X}, \mathbf{X}_*$ is also a Gaussian distribution \cite{rasmussen2006gaussian}:
\begin{equation}
\mathbf{y}_*\vert \mathbf{y}, \mathbf{X}, \mathbf{X}_* \sim \mathcal{N}(\mathbf{\mu}_*,\mathbf{K}_*),
\label{eq:GP_inference}
\end{equation}
where:
$$
\boldsymbol{\mu}_* = \mathbf{K}\left(\mathbf{X}_*,\mathbf{X}\right)\left(\mathbf{K}\left(\mathbf{X},\mathbf{X}\right) + \sigma^2\mathbb{I}\right)^{-1}\mathbf{y}
$$
$$
\mathbf{K}_* = \mathbf{K}\left(\mathbf{X}_*,\mathbf{X}_*\right) + \sigma^2\mathbb{I}-\mathbf{K}\left(\mathbf{X}_*,\mathbf{X} \right)\left(\mathbf{K}\left(\mathbf{X},\mathbf{X}\right) + \sigma^2\mathbb{I}\right)^{-1}\mathbf{K}\left(\mathbf{X},\mathbf{X}_*\right)
$$

\paragraph{Continuous Normalizing Flows.} Normalizing Flows (NF) \cite{rezende2015variational} are gaining popularity among generative models thanks to their flexibility and the ease of training via direct negative log-likelihood (NLL) optimization.
Flexibility is given by the change-of-variable technique that maps a latent variable $\mathbf{z}$ with know prior $p(\mathbf{z})$ to $\mathbf{y}$ from some observed space with unknown distribution. This mapping is performed through a series of (parametric) invertible functions: $\mathbf{y}=\mathbf{f}_n \circ \dots \circ \mathbf{f}_1(\mathbf{z})$. Assuming known prior $p(\mathbf{z})$ for $\mathbf{z}$, the log-likelihood for $\mathbf{y}$ is given by:
\begin{equation}
\log p(\mathbf{y}) = \log p(\mathbf{z}) - \sum_{n=1}^N \log  \left| \det \frac{\partial \mathbf{f}_n}{\partial \mathbf{z}_{n-1}} \right|,
\end{equation}
where $\mathbf{z} = \mathbf{f}_1^{-1}\circ \dots \circ \mathbf{f}_n^{-1} (\mathbf{y})$ is a result of the invertible mapping. The biggest challenge in normalizing flows is the choice of the invertible functions $\mathbf{f}_n, \dots, \mathbf{f}_1$. This is due to the fact that they need to be expressive while guaranteeing an efficient calculation of the Jacobian determinant, which usually has a cubic cost. An alternative approach is given by CNF models \cite{grathwohl2018ffjord}. CNFs use continuous, time-dependent transformations instead of sequence of discrete functions $\mathbf{f}_n, \dots, \mathbf{f}_1$. Formally, we introduce a function $\mathbf{g}_{\boldsymbol{\beta}}(\mathbf{z}(t), t)$ that models the dynamics of $\mathbf{z}(t)$, $\frac{\partial \mathbf{z}(t)}{\partial t}  =\mathbf{g}_{\boldsymbol{\beta}}(\mathbf{z}(t), t)$, parametrized by $\boldsymbol{\beta}$.
In the CNF setting, we aim at finding a solution $\mathbf{y}:=\mathbf{z}(t_1)$ for the differential equation, assuming the given initial state $\mathbf{z}:=\mathbf{z}(t_0)$ with a known prior. As a consequence, the transformation function $\mathbf{f}_{\boldsymbol{\beta}}$ is defined as:
\begin{equation}
\mathbf{y} = \mathbf{f}_{\boldsymbol{\beta}}( \mathbf{z} ) =  \mathbf{z} + \int^{t_1}_{t_0} \mathbf{g}_{\boldsymbol{\beta}}(\mathbf{z}(t), t) dt.
\label{eq:f}
\end{equation}

The inverted form of the transformation can be easily computed using the formula:

\begin{equation}
\mathbf{f}_{\boldsymbol{\beta}}^{-1}( \mathbf{y})  = \mathbf{y} - \int^{t_1}_{t_0} \mathbf{g}_{\boldsymbol{\beta}}(\mathbf{z}(t), t) dt.
\end{equation}

The log-probability of $\mathbf{y}$ can be computed by:
\begin{equation}
\log p(\mathbf{y}) = \log p( \mathbf{f}_{\boldsymbol{\beta}}^{-1}( \mathbf{y}) ) - \int^{t_1}_{t_0} \mathrm{Tr} \left( \frac{\partial \mathbf{g}_{\boldsymbol{\beta}}}{\partial \mathbf{z}(t)} \right) dt
\quad \text{where} \quad 
\mathbf{f}_{\boldsymbol{\beta}}^{-1}( \mathbf{y}) =  \mathbf{z}.
\label{eq:flow_p_y}
\end{equation}

\section{Non-Gaussian Gaussian Processes}

\begin{wrapfigure}{r}{7cm}
\vspace{-0.7cm}
\centering
    \includegraphics[width=0.5\textwidth]{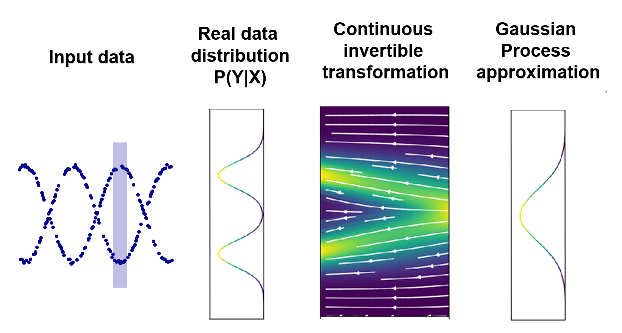}
  \caption{General idea of \our{}. A complex multi-modal distribution can be modelled by exploiting a continuous invertible transformation to fit the Normal distribution used by the GP. Image inspired by Figure~1 in \cite{grathwohl2018ffjord}.
  }
 \label{fig:nggp_diagram}
\end{wrapfigure}

In this work, we introduce Non-Gaussian Gaussian Processes (NGGPs) to cope with the significant bottlenecks of Gaussian Processes for Few-Shot regression tasks: reduced flexibility and assumption about the high similarity between the structure of subsequent tasks. We propose to model the posterior predictive distribution as non-Gaussian on each datapoint. We are doing so by incorporating the flexibility of CNFs. However, we do not stack the CNF on GP to model the multidimensional distribution over $\mathbf{y}$. Instead, we attack the problem with an invertible ODE-based mapping that can utilize each component of the random variable vector and create the specific mapping for each datapoint (see Figure~\ref{fig:nggp_diagram}). 

\begin{figure}[t!]
\centering
    \includegraphics[width=0.90\textwidth]{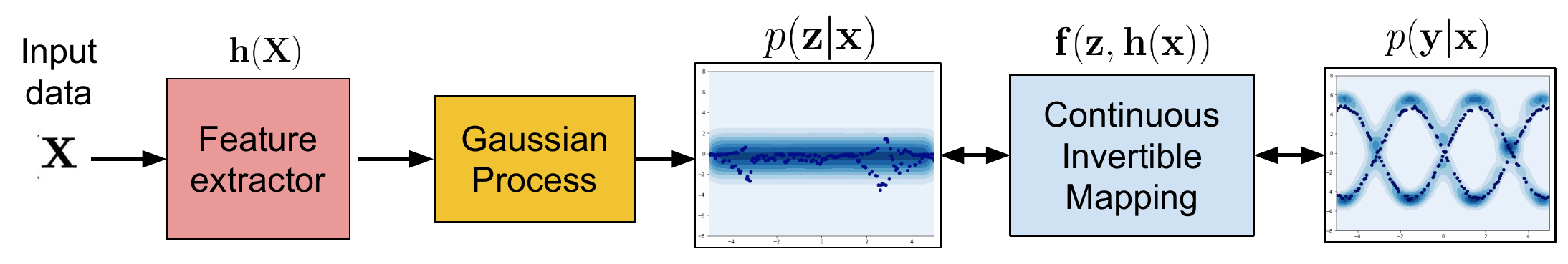}
  \caption{The general architecture of our approach. The input data are embedded by the feature extractor $\mathbf{h(\cdot)}$ and then used to create a kernel for the GP. Next, the output $\mathbf{z}$ of the GP is adjusted using an invertible mapping $\mathbf{f}(\cdot)$ which is conditioned on the output of the feature extractor. This allows us to model complex distributions of the target values $\mathbf{y}$.  }\label{fig:nggp_diagram_2}
\end{figure}

The general overview of our method is presented in Figure~\ref{fig:nggp_diagram_2}. Consider the data matrix $\mathbf{X}$, which stores the observations $\mathbf{x}_i$ for a given task. Each element is processed by a feature extractor $\mathbf{h}(\cdot)$ to create the latent embeddings. Next, we model the distribution of the latent variable $\mathbf{z}$ with a GP. Further, we use an invertible mapping $\mathbf{f}(\cdot)$ in order to model more complex data distributions. Note that the transformation is also conditioned on the output of the feature extractor $\mathbf{h}(\cdot)$ to include additional information about the input.

The rest of this section is organized as follows. In Section~\ref{sec:training}, we demonstrate how the marginal can be calculated during training. In Section \ref{sec:posterior}, we demonstrate how to perform an inference stage with the model. Finally, in Section~\ref{sec:few_shot}, we show how the model is applied to the few-shot setting.

\subsection{Training objective}
\label{sec:training}
Consider the GP with feature extractor $\mathbf{h}_{ \boldsymbol{\phi}}(\cdot)$ parametrized by $\boldsymbol{\phi}$ and any kernel function $k_{\boldsymbol{\theta}}(\cdot,\cdot)$ parametrized by $\boldsymbol{\theta}$. Assuming the given input data $\mathbf{X}$ and corresponding output values $\mathbf{z}$, we can define the marginal log-probability for the GP:
\begin{equation}
\log p(\mathbf{z}|\mathbf{X}, \boldsymbol{\phi}, \boldsymbol{\theta}) = -\frac{1}{2} \mathbf{z}^\mathrm{T}(\mathbf{K} + \sigma^2\mathbb{I})^{-1}\mathbf{z}  -\frac{1}{2} \log |\mathbf{K} + \sigma^2\mathbb{I}| - \frac{D}{2} \log (2 \pi),
\label{eq:marginal_gp}
\end{equation}
where $D$ is the dimension of $\mathbf{y}$, $\mathbf{K}$ is the kernel matrix, and $k_{i,j}=k_{\boldsymbol{\theta}}(\mathbf{h}_{ \boldsymbol{\phi}}(\mathbf{x}_i), \mathbf{h}_{\boldsymbol{\phi}}(\mathbf{x}_j))$.

Taking into account Equation~\eqref{eq:flow_p_y} we can express the log marginal likelihood as follows:
\begin{equation}
\log p(\mathbf{y}|\mathbf{X}, \boldsymbol{\phi}, \boldsymbol{\theta}, \boldsymbol{\beta}) = \log p( \mathbf{z}|\mathbf{X}, \boldsymbol{\phi}, \boldsymbol{\theta}) -  \int^{t_1}_{t_0} \mathrm{Tr} \left( \frac{\partial \mathbf{g}_{\boldsymbol{\beta}}}{\partial \mathbf{z}(t)} \right) dt,
\end{equation}
where $\mathbf{f}_{\boldsymbol{\beta}}^{-1}( \mathbf{y}) =  \mathbf{z}$,  $p( \mathbf{z} |\mathbf{X}, \boldsymbol{\phi}, \boldsymbol{\theta})$ is the marginal defined by Equation~(\ref{eq:marginal_gp}) and $\mathbf{f}_{\boldsymbol{\beta}}^{-1}(\cdot)$ is the transformation given by Equation~(\ref{eq:f}). In the next stage of the pipeline, we propose to apply the flow transformation $\mathbf{f}_{\boldsymbol{\beta}}^{-1}(\cdot)$ independently to each one of the marginal elements in $\mathbf{y}$, that is $\mathbf{f}_{\boldsymbol{\beta}}^{-1}(\mathbf{y})=[f_{\boldsymbol{\beta}}^{-1}(y_1),\dots,f_{\boldsymbol{\beta}}^{-1}(y_D)]^{\mathrm{T}}$, with $f_{\boldsymbol{\beta}}^{-1}(\cdot)$ sharing its parameters across all components. In other words, while the GP captures the dependency across the variables, the flow operates independently on the marginal components of $\mathbf{y}$. Additionally, the flow is conditioned on the information encoded by the feature extractor, such that it can account for the context information $\mathbf{h}_{\boldsymbol{\phi}}(\mathbf{x}_d)$ from the corresponding input value $\mathbf{x}_d$:
\vspace{-0.2cm}
\begin{equation}
y_d = f_{\boldsymbol{\beta}}( z_d, \mathbf{h}_{\boldsymbol{\phi}}(\mathbf{x}_d) ) =  z_d + \int^{t_1}_{t_0} g_{\boldsymbol{\beta}}(z_d(t), t, \mathbf{h}_{\boldsymbol{\phi}}(\mathbf{x}_d)) dt.
\label{eq:f_c}
\end{equation}
The inverse transformation can be easily calculated with the following formula:

\begin{equation}
f_{\boldsymbol{\beta}}^{-1}(y_d) = y_d -  \int^{t_1}_{t_0} g_{\boldsymbol{\beta}}(z_d(t), t, \mathbf{h}_{\boldsymbol{\phi}}(\mathbf{x}_d)) dt
\label{eq:inv_f_c}
\end{equation}

The final marginal log-likelihood can be expressed as:
\vspace{-0.2cm}
\begin{equation}
\log p(\mathbf{y}|\mathbf{X}, \boldsymbol{\phi}, \boldsymbol{\theta}, \boldsymbol{\beta}) = \log p(\mathbf{z}^{\mathbf{h}} |\mathbf{X}, \boldsymbol{\phi}, \boldsymbol{\theta}) - \sum_{d=1}^D \int^{t_1}_{t_0}  \frac{\partial g_{\boldsymbol{\beta}}}{\partial z_d(t)}  dt,
\label{eq:marginal}
\end{equation}
where $\mathbf{z}^{\mathbf{h}}=\mathbf{f}_{\boldsymbol{\beta}}^{-1}( \mathbf{y}, \mathbf{h}_{\boldsymbol{\phi}}(\mathbf{X}))$ is the vector of inverse functions $f_{\boldsymbol{\beta}}( z_d, \mathbf{h}_{\boldsymbol{\phi}}(\mathbf{x}_d) )$ given by  Equation~\eqref{eq:inv_f_c}. 

The transformation described above can be paired with popular CNF models. Here we choose Ffjord~\cite{grathwohl2018ffjord}, which has showed to perform better on low-dimensional data when compared against discrete flows like RealNVP~\cite{dinh2016density} or Glow~\cite{kingma2018glow}. 
Note that, the CNF is applied independently on the components of the GP outputs and shared across them. Therefore, we do not have any issue with the estimation of the Jacobian, since this corresponds to the first-order derivative of the output w.r.t. the scalar input.

\subsection{Inference with the model}
\label{sec:posterior}
At inference time, we estimate the posterior predictive distribution $p(\mathbf{y}_*|\mathbf{X}_*, \mathbf{y}, \mathbf{X}, \boldsymbol{\phi}, \boldsymbol{\theta}, \boldsymbol{\beta})$, where we have access to training data $(\mathbf{y}, \mathbf{X})$ and model the probability of $D_*$ test outputs $\mathbf{y}_*$ given the inputs $\mathbf{X}_*$. The posterior has a closed expression (see Section~\ref{sec:bg}). Since the transformation given by Equation~(\ref{eq:f_c}) operates independently on the outputs, we are still able to model the posterior in closed form: 
\vspace{-0.15cm}
\begin{equation}
\log p(\mathbf{y}_*|\mathbf{X}_*, \mathbf{y}, \mathbf{X}, \boldsymbol{\phi}, \boldsymbol{\theta}, \boldsymbol{\beta}) = \log p(\mathbf{z}_{*}^{\mathbf{h}}|\mathbf{X}, \mathbf{z}^{\mathbf{h}}, \mathbf{X}, \boldsymbol{\phi}, \boldsymbol{\theta}) -  \sum_{d=1}^{D_{*}} \int^{t_1}_{t_0}  \frac{\partial g_{\boldsymbol{\beta}}}{\partial z_{d}(t)}  dt, 
\label{eq:posterior}
\end{equation}
where $\mathbf{z}_{*}^{\mathbf{h}}=f_{\boldsymbol{\beta}}^{-1}( \mathbf{y}_*, \mathbf{h}_{\boldsymbol{\phi}}(\mathbf{X}_*) )$, $\mathbf{z}^{\mathbf{h}}=f_{\boldsymbol{\beta}}^{-1}( \mathbf{y}, \mathbf{h}_{\boldsymbol{\phi}}(\mathbf{X}) )$ are the inverted transformations for test and train data, and $p(\mathbf{z}_{*}^{\mathbf{h}}|\mathbf{X}_*, \mathbf{z}^{\mathbf{h}}, \mathbf{X}, \boldsymbol{\phi}, \boldsymbol{\theta})$ is the GP posterior described in Equation~\eqref{eq:GP_inference}. 

\begin{algorithm}[t]
\small
\caption{\our\ in the few-shot setting, train and test functions.}
\label{alg_overview}
\textbf{Require:}  $\mathcal{D} = \{\mathcal{T}_n\}_{n=1}^{N}$ train dataset and $\mathcal{T}_{*}=\{\mathcal{S}_{*}, \mathcal{Q}_{*}\}$ test task. \\
\textbf{Parameters:}  $\boldsymbol{\theta}$ kernel hyperparameters,  $\boldsymbol{\phi}$ feature extractor parameters, $\boldsymbol{\beta}$ flow transformation parameters.\\ 
\textbf{Hyperparameters:}  $\alpha$, $\eta$, $\gamma$: step size hyperparameters for the optimizers.
\begin{algorithmic}[1]
\vspace{0.1cm}
\Function{Train}{$\mathcal{D}$, $\alpha$, $\eta$, $\gamma$, $\boldsymbol{\theta}$, $\boldsymbol{\phi}$, $\boldsymbol{\beta}$}
\While{not done}
    \State Sample task $\mathcal{T}=(\mathbf{X}, \mathbf{y}) \sim \mathcal{D}$
        \State $\mathcal{L}=-\log p(\mathbf{y} | \mathbf{X}, \boldsymbol{\theta}, \boldsymbol{\phi}, \boldsymbol{\beta})$\Comment{See Equation~\eqref{eq:marginal}}
        \State Update $\boldsymbol{\theta} \leftarrow \boldsymbol{\theta} - \alpha \nabla_{\theta} \mathcal{L}$,\Comment{Updating kernel hyperparameters}
        \State \ \ \ \ \ \ \ \ \ \ \ \  $\boldsymbol{\phi} \leftarrow \boldsymbol{\phi} - \eta \nabla_{\phi} \mathcal{L}$,\Comment{Updating feature extractor parameters}
        \State \ \ \ \ \ \ \ \ \ \ \ \  $\boldsymbol{\beta} \leftarrow \boldsymbol{\beta} - \gamma \nabla_{\beta} \mathcal{L}$\Comment{Updating flow transformation parameters}
\EndWhile
\State \Return $\boldsymbol{\theta}$, $\boldsymbol{\phi}$, $\boldsymbol{\beta}$
\EndFunction
\vspace{0.1cm} 
\Function{Test}{$\mathcal{T}_{*}$, $\boldsymbol{\theta}$, $\boldsymbol{\phi}$, $\boldsymbol{\beta}$}
        \State Assign support $\mathcal{S}_*=(\mathbf{X}_{*,s}, \mathbf{y}_{*,s})$ and query $\mathcal{Q}_*=(\mathbf{X}_{*,q}, \mathbf{y}_{*,q})$
        \State \Return $p(\mathbf{y}_{*,q} | \mathbf{X}_{*,q}, \mathbf{y}_{*, s}, \mathbf{X}_{*, s},, \boldsymbol{\theta}, \boldsymbol{\phi},\boldsymbol{\beta})$ \Comment{See Equation~\eqref{eq:posterior}}
\EndFunction
\Statex
\end{algorithmic}
  \vspace{-0.2cm}
\end{algorithm}
\newpage
\subsection{Adaptation for few-shot regression}
\label{sec:few_shot}

In few-shot learning, we are given a meta-dataset of tasks $\mathcal{D} = \{\mathcal{T}_n\}_{n=1}^{N}$ where each task $\mathcal{T}_n$ contains a support set $\mathcal{S}_n$, and a query set $\mathcal{Q}_n$. At training time, both support and query contain input-output pairs $(\mathbf{X}, \mathbf{y})$, and the model is trained to predict the target in the query set given the support. At evaluation time, we are given a previously unseen task $\mathcal{T}_{\ast}=(\mathcal{S}_{\ast}, \mathcal{Q}_{\ast})$, and the model is used to predict the target values of the unlabeled query points. We are interested in few-shot regression, where inputs are vectors and outputs are scalars. 

We follow the paradigm of Deep Kernel Transfer (DKT) introduced in \cite{patacchiola2020bayesian} and propose the following training and testing procedures (see Algorithm \ref{alg_overview}). During the training stage, we randomly sample the task, calculate the loss defined by  Equation~\eqref{eq:marginal} and update all the parameters using gradient-based optimization. During testing, we simply identify the query and support sets and calculate the posterior given by Equation~\eqref{eq:posterior}. 

\section{Experiments}
\label{sec:experiments}

In this section, we provide an extensive evaluation of our approach (\our) on a set of challenging few-shot regression tasks. We compare the results with other baseline methods used in this domain. As quantitative measures, we use the standard mean squared error (\emph{MSE}) and, when applicable, the negative log-likelihood (\emph{NLL}). 

\begin{figure}[t]
    \centering
    \begin{subfigure}[t]{0.49\textwidth}
        \centering
        \includegraphics[height=4.3cm]{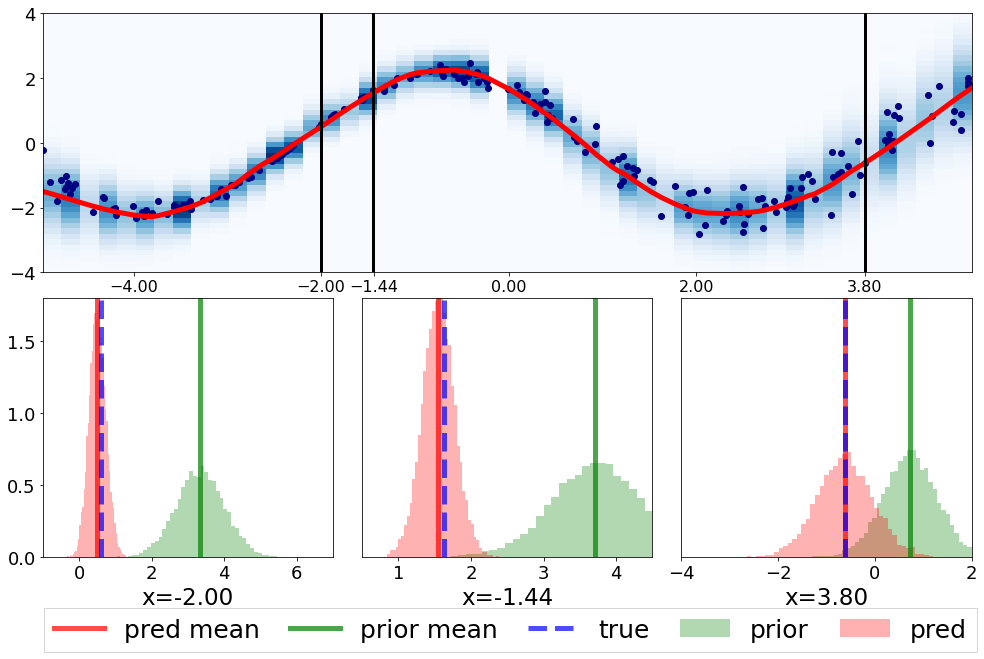}
        \caption{\our\ + NN Linear}
        \label{fig:sines_flow}
    \end{subfigure}%
    \hfill
    \begin{subfigure}[t]{0.49\textwidth}
        \centering
        \includegraphics[height=4.3cm]{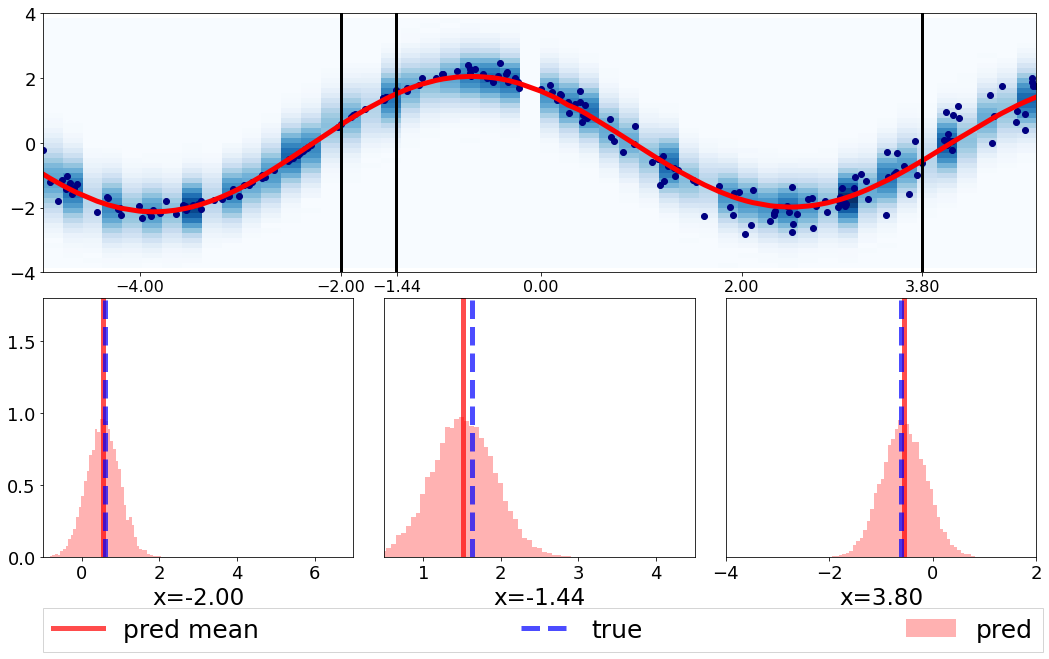}
        \caption{DKT + Spectral}
        \label{fig:sines_dkt}
    \end{subfigure}
    \caption{The results for the sines dataset with mixed-noise for the best performing kernels for \our\ (NN Linear) and DKT (Spectral). The top plot in each figure represents the estimated density (blue hue) and predicted curve (red line), as well as the true test samples (navy blue dots). For three selected input points (denoted by black vertical lines), we plot the obtained marginal densities in the bottom images (red color). In addition, for the \our\ method, we also plot the marginal priors (in green) for each of these three points. It may be observed that \our\ is more successful in modeling the marginal for varying noise levels. }
    \label{fig:sines}
\end{figure}

\paragraph{Sines dataset}
We start by comparing \our\ to other few-shot learning algorithms in a simple regression task defined on sines functions. To this end, we adapt the dataset from \cite{finn2017model} in which every task is composed of points sampled from a sine wave with amplitude in the range $[0.1,5.0]$, phase in the range $[0, \pi]$, and Gaussian noise~$\mathcal{N}(0, 0.1)$. The input points are drawn uniformly at random from the range $[-5,5]$. We consider $5$ support and $5$ query points during the training and $5$ support and $200$ query points during inference. In addition, following \cite{patacchiola2020bayesian}, we also consider an \textit{out-of-range} scenario, in which the range during the inference is extended to $[-5,10]$.
We also perform a variation of sines experiment in which we inject input-dependent noise. The target values in this setting are modeled by $A\sin{(x+\varphi)+|x+\varphi|\epsilon}$, where the amplitude, phase, input, and noise points are drawn from the same distributions as in the standard setup described before. We refer to this dataset ablation as \emph{mixed-noise sines}. For more information about the training regime and architecture, refer to Supplementary Materials~\ref{sec:Appendix_Training}. Table~\ref{tab:sines} presents the results of the experiments. We use the DKT method as a reference since it provides state-of-the-art results for the few-shot sines dataset~\cite{patacchiola2020bayesian}. For a report with more baseline methods, please refer to Supplementary Materials~\ref{sec:Appendix_Sines}.

Both DKT and our \our\  perform very well when paired with the Spectral Mixture Kernel, achieving the same performance on \textit{in-range} data. However, our approach  gives superior results in the \textit{out-of-range} scenario, confirming that \our\ is able to provide a better estimate of the predictive posterior for the unseen portions of the task. It is also worth noting that in all settings, \our\  consistently achieves the best \emph{NLL} results. This is particularly evident for the \textit{in-range} mixed-noise sines dataset. We analyze this result in Figure \ref{fig:sines}, where \our\ successfully models the distribution of the targets, predicting narrow marginals for the more centralized points and using wider distributions for the points with larger noise magnitude. This is in contrast with DKT, which fails to capture different noise levels within the data. These observations confirm our claim that the \our\ is able to provide a good estimate in the case of heteroscedastic data.

\begin{table}[t]
\centering
\caption{The \textit{MSE} and \textit{NLL} results for the inference tasks on sines datasets in the \textit{in-range} and \textit{out-range} settings. Lowest results in bold (the lower the better).}
\label{tab:sines}

\scalebox{0.68}{
\begin{tabular}{lllllllll}
\toprule

\multicolumn{1}{c}{\multirow{3}{*}{\textbf{Method}}}      &
\multicolumn{4}{c}{\textbf{sines}} &
\multicolumn{4}{c}{\textbf{mixed-noise sines}} \\
\cline{2-9} 
\multicolumn{1}{c}{}    &
\multicolumn{2}{c}{\textbf{in-range}}                                  & \multicolumn{2}{c}{\textbf{out-of-range}} &
\multicolumn{2}{c}{\textbf{in-range}} &
\multicolumn{2}{c}{\textbf{out-of-range}}
\\ 
\cline{2-9} 
\multicolumn{1}{c}{}                                      & \multicolumn{1}{c}{\textit{MSE}}  & \multicolumn{1}{c}{\textit{NLL}} & \multicolumn{1}{c}{\textit{MSE}} & \multicolumn{1}{c}{\textit{NLL}} &
\multicolumn{1}{c}{\textit{MSE}} & \multicolumn{1}{c}{\textit{NLL}} &
\multicolumn{1}{c}{\textit{MSE}} & \multicolumn{1}{c}{\textit{NLL}}\\ \hline
\multicolumn{1}{l}{DKT + RBF}                            & \multicolumn{1}{l}{1.36$\pm$1.64} & -0.76$\pm$0.06                     & 2.94$\pm$2.70                     & -0.69$\pm$0.06    
& 1.60$\pm$1.63  & 0.48 $\pm$ 0.22 & 2.99$\pm$ 2.37
& 2.01 $\pm$ 0.59 \\
\multicolumn{1}{l}{DKT + Spectral}                       & \multicolumn{1}{l}{\textbf{0.02$\pm$0.01}} & \textbf{-0.83$\pm$0.03}                     & 0.04$\pm$0.03                     & -0.70$\pm$0.14 &
\textbf{0.18 $\pm$ 0.12} & 0.37$\pm$0.16  &
1.33 $\pm$ 1.10 & 1.58 $\pm$ 0.40 \\
\multicolumn{1}{l}{DKT + NN Linear}                      & \multicolumn{1}{l}{\textbf{0.02$\pm$0.02}} & -0.73$\pm$0.11                    & 6.61$\pm$31.63                     & 38.38$\pm$40.16  &
\textbf{0.18$\pm$0.11} & 0.45 $\pm$ 0.23 &
5.85 $\pm$ 12.10 & 8.64 $\pm$ 6.55 \\
\hline
\multicolumn{1}{l}{\our\ + RBF} & \multicolumn{1}{l}{1.02$\pm$1.40} & -0.74$\pm$0.07                    & 3.02$\pm$2.53                     &-0.65$\pm$0.08  & 1.30$\pm$1.36 & 0.33 $\pm$ 0.16 &
3.90 $\pm$ 2.60 & 1.83 $\pm$ 0.53 \\
\multicolumn{1}{l}{\our\ + Spectral}  & \multicolumn{1}{l}{\textbf{0.02$\pm$0.01}} & \textbf{-0.83$\pm$0.05}                    & \textbf{0.03$\pm$0.02}                     & \textbf{-0.80$\pm$0.07} & 0.22 $\pm$ 0.14 & 0.44 $\pm$ 0.19 &
\textbf{1.14 $\pm$ 0.90} & \textbf{1.35 $\pm$ 0.38} \\
\multicolumn{1}{l}{\our\ + NN Linear}  & \multicolumn{1}{l}{0.04$\pm$0.03} & -0.73$\pm$0.10                    & 7.34$\pm$12.85                     & 29.86$\pm$27.97 & 0.20 $\pm$ 0.12 & \textbf{0.17 $\pm$ 0.15} &
4.74 $\pm$ 6.29 & 2.92 $\pm$ 1.93 \\
\bottomrule
\end{tabular}
}
\end{table}

\paragraph{Head-pose trajectory}

In this experiment, we use the Queen Mary University of London multiview face dataset
\cite{gong1996investigation}. This dataset is composed of grayscale face images of 37 people (32 train, 5 test). There are 133 facial images per person, covering a viewsphere
of $\pm 90^{\circ}$ in yaw and $\pm 30^{\circ}$ in tilt at $10^\circ$ increment.
We follow the evaluation procedure provided in \cite{patacchiola2020bayesian}. Each task consists of randomly sampled trajectories taken from this discrete manifold. The \emph{in-range} scenario includes the full manifold, while the \emph{out-of-range} scenario includes only the leftmost 10 angles. At evaluation time, the inference is performed over the full manifold with the goal of predicting the tilt. The results are provided in Table~\ref{tab:qmul}. In terms of \emph{MSE}, our \our\ method is competitive with other approaches, but it achieves significantly better \emph{NLL} results, especially in the \emph{out-of-range} setting. This suggests that \our s are indeed able to adapt to the differences between the tasks seen at training time and tasks seen at evaluation time by providing a probability distribution that accurately captures the true underlying data.
\begin{table}[h!]
\begin{center}
\begin{scriptsize}
\caption{Quantitative results for Queen Mary University of London for \emph{in-range} and \emph{out-of-range} settings, taking into account \emph{NLL} and \emph{MSE} measures. }
\label{tab:qmul}
\begin{tabular}{lcccc}
\toprule
\multicolumn{1}{c}{\multirow{2}{*}{\textbf{Method}}} & \multicolumn{2}{c}{\textbf{in-range}}                                 & \multicolumn{2}{c}{\textbf{out-of-range}}       \\ \cline{2-5} 
\multicolumn{1}{c}{}                                 & \multicolumn{1}{c}{\textit{MSE}}           & \textit{NLL}            & \textit{MSE}           & \textit{NLL}            \\ \hline
\multicolumn{1}{l}{Feature Transfer/1}              & \multicolumn{1}{c}{0.25$\pm$0.04}          & -                       & 0.20$\pm$0.01          & -                       \\
\multicolumn{1}{l}{Feature Transfer/100}            & \multicolumn{1}{c}{0.22$\pm$0.03}          & -                       & 0.18$\pm$0.01          & -                       \\
\multicolumn{1}{l}{MAML (1 step)}                   & \multicolumn{1}{c}{0.21$\pm$0.01}          & -                       & 0.18$\pm$0.02          & -                       \\
\multicolumn{1}{l}{DKT + RBF}                       & \multicolumn{1}{c}{0.12$\pm$0.04}          & 0.13$\pm$0.14           & 0.14$\pm$0.03          & 0.71$\pm$0.48           \\
\multicolumn{1}{l}{DKT + Spectral}                  & \multicolumn{1}{c}{0.10$\pm$0.01}          & 0.03$\pm$0.13           & 0.07$\pm$0.05          & 0.00$\pm$0.09           \\
\multicolumn{1}{l}{DKT + NN Linear}                 & \multicolumn{1}{c}{0.04$\pm$0.03}          & -0.12$\pm$0.12          & 0.12$\pm$0.05          & 0.30$\pm$0.51           \\
\hline
\multicolumn{1}{l}{\our\ + NN Linear} & \multicolumn{1}{c}{0.02$\pm$0.02}          & -0.47$\pm$0.32          & 0.06$\pm$0.05          & 0.24$\pm$0.91           \\
\multicolumn{1}{l}{\our\ + Spectral}  & \multicolumn{1}{c}{\textbf{0.03$\pm$0.03}} & \textbf{-0.68$\pm$0.23} & \textbf{0.03$\pm$0.03} & \textbf{-0.62$\pm$0.24} \\
\bottomrule
\end{tabular}
\end{scriptsize}
\end{center}

\end{table}

\paragraph{Object pose prediction}
We also study the behavior of \our\ in a pose prediction dataset introduced in~\cite{yin2019meta}. Each task in this dataset consists of $30$ gray-scale images with resolution $128\times128$, divided evenly into support and query. The tasks are created by selecting an object from the Pascal 3D~\cite{xiang2014beyond} dataset, rendering it in $100$ random orientations, and sampling out of it $30$ representations. The goal is to predict the orientation relative to a fixed canonical pose. Note that $50$ randomly selected objects are used to create the meta-training dataset, while the remaining $15$ are utilized to create a distinct meta-test set. Since the number of objects in meta-training is small, a model could memorize the canonical pose of each object and then use it to predict the target value, completely disregarding the support points during the inference. This would lead to poor performance on the unseen objects in the meta-test tasks. This special case of overfitting is known as the \textit{memorization problem}~\cite{yin2019meta}. 

We analyze the performance of GP-based models in this setting by evaluating the performance of DKT and \our\  models\footnote{Information about architecture and training regime is given in Supplementary Materials \ref{sec:Appendix_Training}.}. We compare them against the methods used in~\cite{yin2019meta}, namely  MAML~\cite{finn2017model}, Conditional Neural Processes (CNP)~\cite{garnelo2018neural} and their meta-regularized versions devised to address the memorization problem --- MR-MAML and MR-CNP~\cite{yin2019meta}. In addition, we also include the fine-tuning (FT) baseline and CNP versions with standard regularization techniques such as Bayes-by-Backprop (BbB)~\cite{blundell2015weight} and Weight Decay~\cite{krogh1992simple}. The results are presented in Table~\ref{object_pose_table}.

\begin{wraptable}{r}{6cm}
  \vspace{-0.4cm}
  \caption{Quantitative results for the object pose prediction task. We report the mean and standard deviation over 5 trials. The lower the better. Asterisks (*) denote values reported in \cite{yin2019meta}.}
  \label{object_pose_table}
  \centering
  \begin{scriptsize}
  \begin{tabular}{lcc}
    \toprule
    \textbf{Method}     & \textit{MSE}     & \textit{NLL} \\
    \midrule
    MAML* & 5.39 $\pm$ 1.31 & - \\
    MR-MAML*  & 2.26 $\pm$ 0.09 & - \\
    CNP* & 8.48 $\pm$ 0.12 & - \\
    MR-CNP* &  2.89 $\pm$ 0.18 & -  \\
    FT* &  7.33 $\pm$ 0.35 & - \\
    FT + Weight Decay* & 6.16 $\pm$ 0.12 & - \\
    \midrule
    CNP + Weight Decay* & 6.86 $\pm$ 0.27 & - \\
    CNP + BbB* & 7.73 $\pm$ 0.82 & - \\
    \midrule
    DKT + RBF &  1.82 $\pm$ 0.17 & 1.35 $\pm$ 0.10\\
    DKT + Spectral & \textbf{1.79} $\pm$ \textbf{0.15} & 1.30 $\pm$ 0.06\\
    \midrule
    \our\ + RBF & 1.98 $\pm$ 0.27 & \textbf{0.22} $\pm$ \textbf{0.08}\\
    \our\ + Spectral &  2.34 $\pm$ 0.28 & 0.86 $\pm$ 0.45\\
    \bottomrule
  \end{tabular}
    \end{scriptsize}
\end{wraptable}

Both GP-related approaches: \our\ and DKT are similar or usually outperform the standard and meta-regularized methods, which indicates that they are less prone to memorization and therefore benefit from a better generalization. The \textit{NLL} is significantly lower for \our\ than for DKT, confirming that \our\ is better at inferring complex data distributions. 

\paragraph{Power Dataset} \label{sec:power}
In this series  of experiments, we use the Power~\cite{Power} dataset and define an experimental setting for the few-shot setting. We treat each time series composed of 1440 values (60 minutes $\times$ 24 hours) that represents the daily power consumption (\emph{sub\_metering\_3}) as a single task. We train the model using the tasks from the first $50$ days, randomly sampling $10$ points per task, while validation tasks are generated by randomly selecting from the following $50$ days. 

\begin{figure}[t]
     \centering
     \begin{subfigure}[b]{0.49\textwidth}
         \centering
         \includegraphics[height=5.5cm]{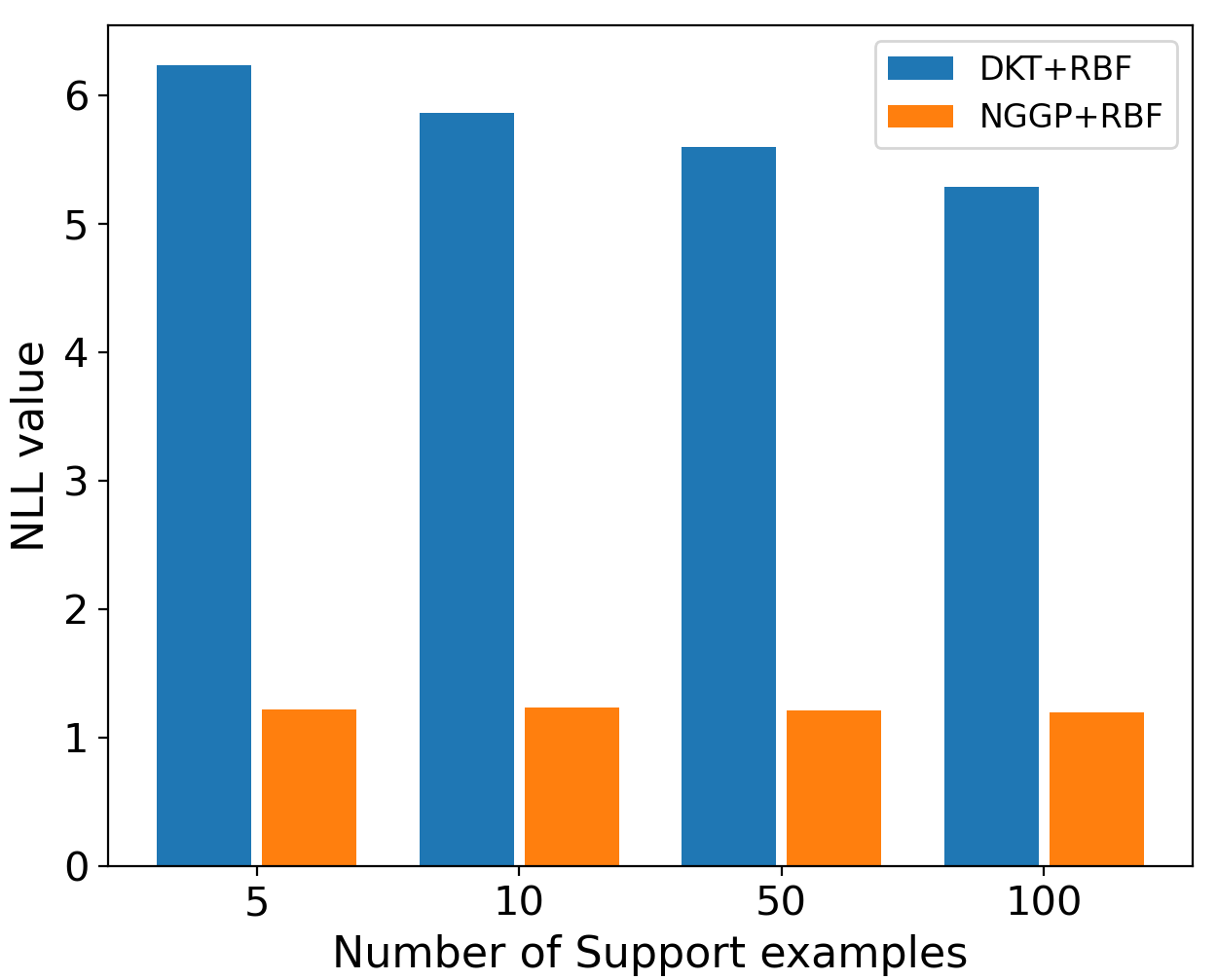}
         \caption{NLL results DKT vs. NGGP.}
         \label{fig:power}
     \end{subfigure}
     \begin{subfigure}[b]{0.49\textwidth}
         \centering
         \includegraphics[height=5.5cm]{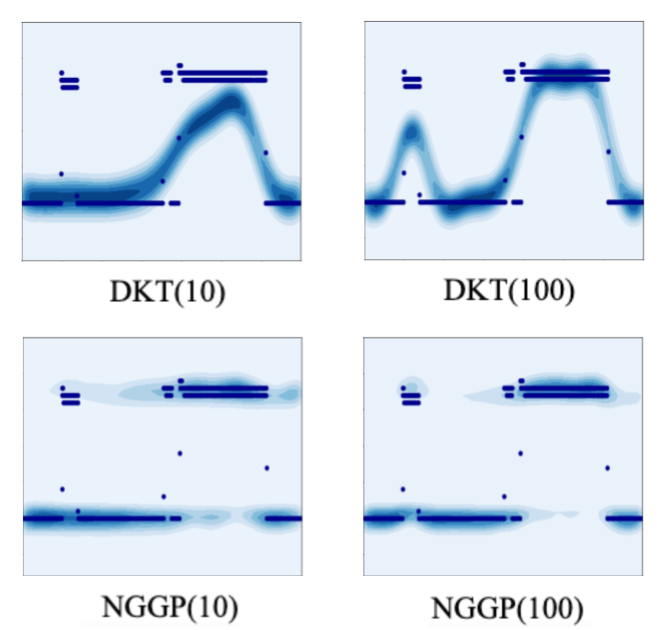}
         \caption{Single day comparison DKT vs. NGGP.}
         \label{fig:power_plots}
     \end{subfigure}
     
        \caption{The results for the Power dataset experiment: \textbf{(a)} The quantitative comparison between DKT and \our\ considering different numbers of support examples. \textbf{(b)} The power consumption for a single day randomly selected from the test data. We compare DKT vs. \our\ (with RBF kernel) considering 10 and 100 support points. \our\ captures multi-modality and thus better adjusts to the data distribution.}
        \label{fig:power_joined}
\end{figure}

Quantitative and qualitative analysis are provided in Figure~\ref{fig:power_joined}. We use only \emph{NLL} to assess the results due to the multi-modal nature of the data and analyze the value of the criterion for different numbers of support examples. \our\ better adjusts to the true data distribution, even in the presence of very few support examples during inference. This experiment supports the claim that \our s are well-suited for modeling multi-modal distributions and step functions.  

\paragraph{NASDAQ and EEG datasets} \label{sec:nasdaq_eeg}
In order to test the performance of our methods for real-world time series prediction, we used two datasets - NASDAQ100~\cite{nasdaq_dataset} and EEG~\cite{eeg_dataset}. For an extensive description of the datasets and evaluation regime of this experiment, see~Supplementary~Materials~\ref{sec:Appendix_Training}. Quantitative results are presented in Table~\ref{tab:nasdaq_eeg}. Our experiments show that \our{} outperforms the baseline DKT method across all datasets. The improvement is especially visible for the \textit{out-of-range} NASDAQ100 when both methods use the RBF kernel. The results suggest that \our s can be successfully used to model real-world datasets, even when the data does not follow a Gaussian distribution.

\begin{table}[ht]
     \caption{Quantitative results for NASDAQ and EEG datasets.}
     \label{tab:nasdaq_eeg}
\scriptsize
    \begin{subtable}[h]{0.495\textwidth}
        \centering
               \caption{NASDAQ100}

\begin{tabular}{lll}
\toprule
\multicolumn{3}{c}{\textbf{in-range}}                                                            \\
\midrule
\multicolumn{1}{c}{\textbf{Method}} & \multicolumn{1}{c}{\textit{MSE $\cdot$ 100}} & \multicolumn{1}{c}{\textit{NLL}} \\
\midrule
\our\ + RBF                          & \textbf{0.012} $\pm$ \textbf{0.014}    & \textbf{-3.092} $\pm$ \textbf{0.255}     \\
\our\ + NN Linear                    & 0.023 $\pm$ 0.044             & -2.567 $\pm$ 1.235              \\
DKT + NN Linear                     & 0.027 $\pm$ 0.032             & -2.429 $\pm$ 0.271              \\
DKT + RBF                           & 0.022 $\pm$ 0.042             & -2.878 $\pm$ 0.706              \\
\toprule
\multicolumn{3}{c}{\textbf{out-of-range}}                                                        \\
\midrule
\multicolumn{1}{c}{\textbf{Method}} & \multicolumn{1}{c}{\textit{MSE $\cdot$ 100}} & \multicolumn{1}{c}{\textit{NLL}} \\
\midrule
\our\ + RBF                          & 0.016 $\pm$ 0.034               & -2.978 $\pm$ 0.571              \\
\our\ + NN Linear                    & \textbf{0.003} $\pm$ \textbf{0.004}      & \textbf{-2.998} $\pm$ \textbf{0.260}     \\
DKT + NN Linear                     & 0.005 $\pm$ 0.006               & -2.612 $\pm$ 0.059             \\
DKT + RBF                           & 0.181 $\pm$ 0.089               & 1.049 $\pm$ 2.028   \\
\bottomrule
\end{tabular}
       \label{tab:week1}
    \end{subtable}
    \hfill
    \begin{subtable}[h]{0.495\textwidth}
        \centering
            \caption{EEG}

\begin{tabular}{lll}
\toprule
\multicolumn{3}{c}{\textbf{in-range}}                                                               \\
\midrule
\multicolumn{1}{c}{\textbf{Method}} & \multicolumn{1}{c}{\textit{MSE $\cdot$ 100}} & \multicolumn{1}{c}{\textit{NLL}} \\
\midrule

\our\ + RBF                          & \textbf{0.222} $\pm$ \textbf{0.181}    & \textbf{-1.715} $\pm$ \textbf{0.282}       \\
\our\ + NN Linear                    & 0.361 $\pm$ 0.223             & -1.387 $\pm$ 0.273                \\
DKT + NN Linear                     & 0.288 $\pm$ 0.169             & -1.443 $\pm$ 0.188                \\
DKT + RBF                           & 0.258 $\pm$ 0.218             & -1.640 $\pm$ 0.237                \\

\toprule

\multicolumn{3}{c}{\textbf{out-of-range}}                                                  \\
\midrule

\multicolumn{1}{c}{\textbf{Method}} & \multicolumn{1}{c}{\textit{MSE $\cdot$ 100}} & \multicolumn{1}{c}{\textit{NLL}} \\
\midrule

\our\ + RBF                          & 0.463 $\pm$ 0.415               & \textbf{-1.447} $\pm$ \textbf{0.221}       \\
\our\ + NN Linear                    & \textbf{0.452} $\pm$ \textbf{0.578}      & -1.046 $\pm$ 0.624                \\
DKT + NN Linear                     & 0.528 $\pm$ 0.642               & -1.270 $\pm$ 0.622                \\
DKT + RBF                           & 0.941 $\pm$ 0.917               & -1.242 $\pm$ 0.685  \\
\bottomrule             
\end{tabular}
        \label{tab:eeg}
     \end{subtable}
\end{table}

\section{Conclusions}\label{sec:conclusions}
In this work, we introduced \our{} -- a generalized probabilistic framework that addresses the main limitations of Gaussian Processes, namely its rigidity in modeling complex distributions.
\our{} leverages the flexibility of Normalizing Flows to modulate the posterior predictive distribution of GPs. Our approach offers a robust solution for few-shot regression since it finds a shared set of parameters between consecutive tasks while being adaptable to dissimilarities and domain shifts.
We have provided an extensive empirical validation of our method, verifying that it can obtain state-of-the-art performance on a wide range of challenging datasets. 
In future work, we will focus on applications of few-shot regression problems needing the estimation of exact probability distribution (\emph{e.g.},  continuous object-tracking) and settings where there is a potential discontinuity in similarity for subsequent tasks (\emph{e.g.}, continual learning).

\paragraph{Limitations}
The main limitation of \our{} s is the costs of learning flow-based models, that could be more expensive than using a standard DKT when the data come from a simple distribution. In such a case, other methods like DKT could be more efficient. Moreover, GPs are expensive for tasks with a large number of observations, making \our{} a better fit for few-shot learning rather than bigger settings. 
Finally, in some cases, it can be more challenging to train and fine-tune \our{} than DKT because the number of parameters and hyper-parameters is overall larger (e.g. the parameters of the flow).

\paragraph{Broader Impact}
Gaussian Processes for regression already have had a huge impact on various real-world applications~\cite{dutordoir2018gaussian, yeung2009learning, NIPS2006_f42c7f9c, li2019kernel}. \our s make it possible to apply \textit{a priori} knowledge and expertise to even more complex real-world systems, providing fair and human-conscious solutions, \emph{i.e.}, in~neuroscience or social studies (see experiments on individual power consumption, EEG, and NASDAQ datasets from section \ref{sec:nasdaq_eeg}). The proposed method is efficient and represents a great tool for better uncertainty quantification. Careful consideration of possible applications of our method must be taken into account to minimize any possible societal impact. For instance, the use of \our{} in object-tracking could be harmful if deployed with malevolent and unethical intents in applications involving mass surveillance. 

\section*{Acknowledgments}

This research was funded by Foundation for Polish Science (grant no POIR.04.04\allowbreak.00-00-14DE/18-00 carried out within the Team-Net program co-financed by the European Union under the European Regional Development Fund) and National Science Centre, Poland (grant no 2020/39/B/ST6/01511). The work of M. Zieba was supported by the National Centre of Science (Poland) Grant No. 2020/37/B/ST6/03463. The work of P. Spurek was supported by the National Centre of Science (Poland) Grant No. 2019/33/B/ST6/00894.
This research was funded by the Priority Research Area Digiworld under
the program Excellence Initiative – Research University at the Jagiellonian
University in Kraków. The authors have applied a CC BY license to any Author Accepted Manuscript (AAM) version arising from this submission, in accordance with the grants’ open access conditions. 
\newpage
{
\small

\bibliographystyle{plain}
\bibliography{references}

}

\newpage
\appendix

\section{Training Regime}
\label{sec:Appendix_Training}

\subsection{Implementation of the GPs}
We use the {\tt GPyTorch}\footnote{https://gpytorch.ai/, available on the MIT Licence} package for the computations of GPs and their kernels. The NN linear kernel is implemented in all experiments as a $1$-layer MLP with ReLU activations and hidden dimension $16$.
For the Spectral Mixture Kernel, we use $4$ mixtures. 

\subsection{Sines Dataset}

For the first experiments on sines functions, we use the dataset from \cite{finn2017model}. For each task, the input points $x$ are sampled from the range $[-5,5]$, and the target values $y$ are obtained by applying $y=A\sin{(x-\varphi)}+\epsilon$, where the amplitude $A$ and phase $\varphi$ are drawn uniformly at random from ranges $[0.1,5]$ and $[0, \pi]$, respectively. The noise values $\epsilon$ are modeled by a normal distribution with zero mean and standard deviation equal to $0.1$. 

During the training, we use $5$ support and $5$ query points. The inference is performed over $500$ tasks, each consisting of $200$ query points and $5$ support points. The models are trained for $50000$ iterations with batch size $1$ (one task per each parameters update) and learning rate $0.001$ using the Adam optimizer with $\beta_1=0.9$ and $\beta_2=0.999$. 

The feature extractor for this experiment is implemented by a $2$-layer MLP with ReLU activations and hidden dimension $40$, which follows the setting of \cite{finn2017model}. The last hidden layer is used as the representation for the DKT\footnote{For the DKT implementation we use the code provided at \url{https://github.com/BayesWatch/deep-kernel-transfer}} and \our\ methods in the Gaussian Process framework. 

The CNF component for our model was inspired by FFJORD. Our implementation is based on the original code provided by the authors\footnote{\url{https://github.com/rtqichen/ffjord}}. We use two stacked blocks of CNFs, each composed of two hidden \emph{concatsquash} layers, $64$ units each, with \emph{tanh} activation. We adjusted \emph{concatsquash} layers for the conditional variant of CNF by feeding them with an additional conditioning factor - the $40$ dim output from the feature extractor.  

We use the same settings for the \textit{in-range} heterogeneous noise experiment, but we train the \our\ method for $10000$ iterations instead of $50000$ since we have noticed that this is enough for the model to converge. 

\subsection{Head-pose trajectory}
For the head-pose trajectory task, we use the same setting as proposed in \cite{patacchiola2020bayesian} with the same feature extractor - convolution neural network with $3$ layers, each with
36 output channels, stride 2, and dilation 2. The NN Linear kernel in this experiment is implemented by a $1$-layer MLP with ReLU activations and hidden dimension $16$.

During the training phase, we use a meta-batch size equal to $5$, the learning rate $0.001$, and the Adam optimizer with the same configuration as in the sines experiment. Models were trained for $100$ iterations. We use $5$ support and $5$ query points during the train. During the inference, we use  $5$ points as the support and the remaining samples of the trajectory as the query. We perform the inference over $10$ different tasks.

For \our{}, we use the same CNF component architecture as in for the sines dataset. However, we also add Gaussian noise from the Normal distribution $\mathcal{N}(0,0.1)$ to the head-pose orientations. Adding noise allows for better performance when learning with the CNF component.

\subsection{Object pose prediction}
In order to verify the extend of memorization in \our, we consider so-called \textit{non-mutually exclusive} tasks. In this setting, the tasks are constructed in such a way that a single model can solve all tasks zero-shot. 
In particular, we follow the procedure of the pose prediction task introduced in \cite{yin2019meta}. 
The few-shot regression dataset is based on the Pascal 3D\footnote{\url{ftp://cs.stanford.edu/cs/cvgl/PASCAL3D+_release1.1.zip}} data \cite{xiang2014beyond} and was recreated based on the code from the original research paper \footnote{\url{https://github.com/google-research/google-research/tree/master/meta_learning_without_memorization}, on  Apache-2.0 License}. Firstly, the objects were randomly split into the meta-training set ($50$) and meta-testing ($15$), then the MuJoCo \cite{todorov2012mujoco} library was used to render the instances of objects on a table, setting them random orientations. The observation is a tuple consisting of a $128 \times 128$ gray-scale image and its label - orientation relative to a fixed canonical pose. Every task consists of $30$ positions sampled from the $100$ renderings and divided randomly into \textit{support} and \textit{query}.

During the training, we use a meta-batch of $10$ tasks. The \our{} and DKT models were trained over $1000$ iterations, with learning rates equal to $0.01$ for the kernel parameters, $0.01$ for the feature extractor parameters, and $0.001$ for the ODE-mapping component. We used the Adam optimizer with the same $\beta$ configuration as in the sines experiment. We also use the same CNF component architecture as in the sines dataset. Similarly, as in the head-pose trajectory experiment, we add Gaussian noise from $\mathcal{N}(0,0.1)$ to the orientations for better performance . The inference is performed over $100$ tasks, which also consist of $15$ support and $15$ query points.
As the feature extractor, we use one of the architectures tested in the original research paper \cite{yin2019meta} - the convolutional encoder with five layers stacked as follows: $2$ convolutional layers with stride $2$ and output dimensions $32$ and $48$; \textit{max pooling layer} with kernel $2 \times 2$; convolutional layer with output dimension $64$; \textit{flatten layer} and \textit{linear layer} with output dimension equal to $64$.

For this dataset, we tested \our{} and DKT models with RBF and Spectral kernels only. This choice was due to the similarity between head-pose trajectory and object pose prediction settings, and the results show that these two kernels performed the best on such tasks.  

\subsection{Power Dataset}

The Power Dataset\footnote{\url{https://archive.ics.uci.edu/ml/datasets/individual+household+electric+power+consumption}, made available under the “Creative Commons Attribution 4.0 International (CC BY 4.0)” license.} is an UCI benchmark that describes individual household electric power consumption. The original data is composed of 7 time-dependent attributes, but we focus only on the \emph{sub\_metering\_3} attribute in our experiments. We split the dataset into tasks, where each of the tasks corresponds to daily electricity consumption and is represented by $1440$ measurements (in minutes). We train the model using the first $50$ days and validate it using the next $50$ days. We used the same architecture as for the sines dataset in our experiments, except the feature extractor returns $1D$ embedding. 

\subsection{NASDAQ100 and EEG Datasets}

The NASDAQ100\footnote{\url{https://cseweb.ucsd.edu/~yaq007/NASDAQ100_stock_data.html}} dataset consists of $81$ major stocks under the NASDAQ 100 index. We decided to use the NASDAQ100 dataset with padding that includes $390$ points per day over a $105$ days interval. 

We use $70\%$ of the initial data points of the NDX100 index for the creation of meta-train tasks.  The \textit{in-range} meta-tasks were obtained from the last $30\%$ of the data, while the \textit{out-of-range} inference was obtained from the whole time-series of a different stock index. For this purpose, we utilize the time-series given by the YHOO index, which was not used during the training.

The EEG\footnote{\url{https://archive.ics.uci.edu/ml/datasets/EEG+Steady-State+Visual+Evoked+Potential+Signals}, UCI repository dataset}
dataset contains raw time series of brainwave signals sampled at 128Hz for $14$ electrodes placed at different areas of the patient scalp. Particular patients had been stimulated for various periods, so the time series had different lengths. 

The meta-training tasks were obtained form patient $A001SB1\_1$ and electrode $AF4$ from the first $70\%$ of that time-series data points. Same as in NASDAQ100, meta-test tasks were for the \textit{in-range} scenario were obtained from the last $30\%$ of the same data. The \textit{out-of-range} inference tasks were computed on different patient time-series of EEG data points - we used the $A003SB1\_1$ patient.

For both models, we used the same backbone architecture with Adam optimizer parameters set to the same values as in the experiment on the sines dataset with a learning rate set~to~$0.001$. During the training and testing, we used $5$ support and $5$ query points. The support and query points where sampled as an random interval of $10$ consecutive points. Models were trained with a batch size $1$ for $1000$ iterations.

\section{Additional Results: Sines Regression}
\label{sec:Appendix_Sines}

In addition to the GP-based methods reported in the main text, we also summarize the performance of other baseline algorithms on the sines dataset with standard Gaussian noise. The results are presented in Table~\ref{tab:Appendix_sines}. It may be observed that the DKT and \our\  significantly outperform other approaches. Therefore we only provide a comparison between those two methods in section \ref{sec:experiments} in the main paper. 

\begin{table}[h!]
\centering
\caption{The \textit{MSE} and \textit{NLL} results for the inference tasks on sines datasets in the \textit{in-range} and \textit{out-range} settings. The lowest results in bold. Asterisks (*) and (**) denote values reported in \cite{tossou2019adaptive} and \cite{patacchiola2020bayesian}, respectively. The lower the result, the better.}
\label{tab:Appendix_sines}
\begin{tabular}{lll|l|l}
\hline

\multicolumn{1}{c}{\multirow{2}{*}{\textbf{Method}}}      & \multicolumn{2}{c|}{\textbf{in-range}}                                  & \multicolumn{2}{c}{\textbf{out-of-range}}                           \\ \cline{2-5} 
\multicolumn{1}{c}{}                                      & \multicolumn{1}{c|}{\textit{MSE}}  & \multicolumn{1}{c|}{\textit{NLL}} & \multicolumn{1}{c|}{\textit{MSE}} & \multicolumn{1}{c}{\textit{NLL}} \\ \hline

\multicolumn{1}{l|}{ADKL*}                   & \multicolumn{1}{l|}{0.14} & -                                 & -                     & -                                \\
\multicolumn{1}{l|}{R2-D2*}                   & \multicolumn{1}{l|}{0.46} & -                                 & -                   & -                                \\
\multicolumn{1}{l|}{ALPaCA**}                   & \multicolumn{1}{l|}{0.14$\pm$0.09} & -                                 & 5.92$\pm$0.11                     & -                                \\

\multicolumn{1}{l|}{Feature Transfer/1**}                   & \multicolumn{1}{l|}{2.94$\pm$0.16} & -                                 & 6.13$\pm$0.76                     & -                                \\
\multicolumn{1}{l|}{Feature Transfer/100**}                 & \multicolumn{1}{l|}{2.67$\pm$0.15} & -                                 & 6.94$\pm$0.97                     & -                                \\
\multicolumn{1}{l|}{MAML (1 step)**}                        & \multicolumn{1}{l|}{2.76$\pm$0.06} & -                                 & 8.45$\pm$0.25                     & -                                \\
\multicolumn{1}{l|}{DKT + RBF}                            & \multicolumn{1}{l|}{1.36$\pm$1.64} & -0.76$\pm$0.06                     & 2.94$\pm$2.70                     & -0.69$\pm$0.06                    \\
\multicolumn{1}{l|}{DKT + Spectral}                       & \multicolumn{1}{l|}{\textbf{0.02$\pm$0.01}} & \textbf{-0.83$\pm$0.03}                     & 0.04$\pm$0.03                     & -0.70$\pm$0.14                    \\
\multicolumn{1}{l|}{DKT + NN Linear}                      & \multicolumn{1}{l|}{0.02$\pm$0.02} & -0.73$\pm$0.11                    & 6.61$\pm$31.63                     & 38.38$\pm$40.16                    \\
\multicolumn{1}{l|}{\our\ + RBF} & \multicolumn{1}{l|}{1.02$\pm$1.40} & -0.74$\pm$0.07                    & 3.02$\pm$2.53                     &-0.65$\pm$0.08                    \\
\multicolumn{1}{l|}{\our\ + Spectral}  & \multicolumn{1}{l|}{\textbf{0.02$\pm$0.01}} & -0.83$\pm$0.05                    & \textbf{0.03$\pm$0.02}                     & \textbf{-0.80$\pm$0.07}  \\
\multicolumn{1}{l|}{\our\ + NN Linear}  & \multicolumn{1}{l|}{0.04$\pm$0.03} & -0.73$\pm$0.10                    & 7.34$\pm$12.85                     & 29.86$\pm$27.97  \\
\bottomrule
\end{tabular}
\end{table}

\section{Additional Results: Classical Regression Tasks}
\label{sec:classsic_regression}

Our main goal was to show improvement of \our\ over standard GPs in the case of a few-shot regression task. Albeit, we test our method also in classical regression task setting. Intuition is that \our\ may be superior to standard GPs in a simple regression setting for datasets with non-gaussian characteristics, but do not expect any improvement otherwise.

\subsection{Classical Regression Tasks}
Following the experiments from \cite{NIPS2012_d840cc5d,NIPS2003_6b5754d7}, we decided to run NGGP on regular regression tasks.  In this setting, we trained models over 10000 iterations on samples containing 100 points from a given dataset. Averaged results on 500 test samples containing 40 points that were not seen during the training - are presented in \ref{tab:Appendix_classical_regression}.

\begin{table}[h!]\scriptsize
\centering
\caption{Results on classical regression tasks on proposed datasets are inconclusive. One may see that results of methods performance vary between datasets.} \label{tab:Appendix_classical_regression}
\begin{tabular}{l|c|c|c|c|c|c}
\toprule
\multicolumn{1}{c|}{\multirow{2}{*}{\textbf{Dataset}}}    & \multicolumn{2}{c|}{\textit{abalone}} & \multicolumn{2}{c|}{\textit{Ailerons}} & \multicolumn{2}{c}{\textit{creeprupt}} \\
\cline{2-7} 
   & MSE       & NLL       & MSE       & NLL       & MSE       & NLL   
   \\
   \midrule
GP + RBF   &   1.26 $\pm$ 0.68        &    \textbf{-1.47 $\pm$ 0.20}      &    1.28 $\pm$ 0.66       &    -1.47 $\pm$ 0.19      &    1.26 $\pm$ 0.68      &   \textbf{-1.47 $\pm$ 0.19}        \\
DKT + RBF  &     \textbf{1.18 $\pm$ 0.28}      &    -1.41 $\pm$ 0.09     &  1.41 $\pm$ 0.39      &  \textbf{-1.49 $\pm$ 0.12}        &   1.24 $\pm$ 0.39      & -1.44 $\pm$ 0.12 \\
NGGP + RBF &    1.23 $\pm$ 0.29       &  -1.44 $\pm$ 0.09       &   \textbf{1.25 $\pm$ 0.31}      &  -1.44 $\pm$ 0.10        &    \textbf{1.10 $\pm$ 0.25}      & -1.41 $\pm$ 0.08          
\end{tabular}
\end{table}

\subsection{Sines}

\begin{table}[h!]
\centering
\caption{One may observe that addition of CNF significantly improves results of the classical GP with RBF kernel in such setting.} \label{tab:Appendix_sines2d}
\begin{tabular}{c|c|c|c}
    & GP + RBF      & DKT + RBF & NGGP + RBF \\ \hline
MSE & 1.06 $\pm$ 0.24  & 0.72 $\pm$ 0.32                 & \textbf{0.34 $\pm$ 0.22}                       \\ \hline
NLL & -0.98 $\pm$ 0.10 & -1.20 $\pm$ 0.15                & \textbf{-1.33 $\pm$ 0.13}                      \\ 
\end{tabular}
\end{table}

We ran additional experiments on a synthetic dataset of 2d sine waves (as in the setting from Figure~\ref{fig:tezer}). The data was generated by randomly sampling either $\sin(x)$ or $-\sin(x)$  for a given point $x$, together with adding uniform noise from $(0.1, 0.5)$. Models were trained for $10000$ iterations over samples from the range $(-5.0, 5.0)$ with $100$ points in one sample.
The prediction was done for samples from the interval $(5.0, 10.0)$ - MSE and NLL were averaged on $500$ test samples. We present the quantitative results in Table \ref{tab:Appendix_sines2d}.

\end{document}